\documentclass[10pt,twocolumn,letterpaper]{article}

\usepackage{cvpr}
\usepackage{times}
\usepackage{epsfig}
\usepackage{graphicx}
\usepackage{amsmath}
\usepackage{amssymb}

\usepackage{epstopdf}
\usepackage[boxed,commentsnumbered,ruled,vlined]{algorithm2e}

\usepackage{paralist}

\cvprfinalcopy 

\def\cvprPaperID{****} 

\begin{document}

\title{Fully-adaptive Feature Sharing in Multi-Task Networks with Applications in Person Attribute Classification}

\author{Yongxi Lu\\
UC San Diego\\
{\tt\small yol070@ucsd.edu}
\and
Abhishek Kumar\\
IBM Research\\
{\tt\small abhishk@us.ibm.com}
\and
Shuangfei Zhai \\
Binghamton Univeristy, SUNY\\
{\tt\small szhai2@binghamton.edu}
\and
Yu Cheng\\
IBM Research \\
{\tt\small chengyu@us.ibm.com}
\and
Tara Javidi\\
UC San Diego\\
{\tt\small tjavidi@eng.ucsd.edu}
\and
Rogerio Feris\\
IBM Research\\
{\tt\small rsferis@us.ibm.com}
}

\maketitle

\vspace{-0.2in}
\begin{abstract}
	Multi-task learning aims to improve generalization performance of multiple prediction tasks
	by appropriately sharing relevant information across them.
	In the context of deep neural networks,
	this idea is often realized by hand-designed
	network architectures with layers that are shared across tasks and branches that encode task-specific features.
	However, the space of possible multi-task deep architectures is combinatorially large
	and often the final architecture is arrived at by manual exploration of this space
	subject to designer's bias, which can be both error-prone and tedious.
	In this work, we propose a principled approach for designing compact multi-task deep learning
	architectures.
	Our approach starts with a thin network and dynamically widens it in a greedy manner during training using a novel
	criterion that promotes grouping of similar tasks together.
	Extensive
	evaluation on person attributes classification tasks involving facial and clothing attributes
	suggests that the models produced by the proposed method are fast, compact and can closely match or exceed the
	state-of-the-art accuracy from strong baselines by much more expensive models. 
	
\end{abstract}

\vspace{-0.2in}
\section{Introduction}


Humans possess a natural yet remarkable ability of seamlessly transferring and sharing knowledge 
across multiple related domains while doing inference for a given task. Effective mechanisms 
for sharing \emph{relevant} information across multiple prediction tasks (referred as \emph{multi-task learning}) 
are also arguably crucial for making significant advances towards machine intelligence. 
In this paper, we
propose a novel approach for multi-task learning in the context of deep neural networks for computer vision tasks. 
We particularly aim for two desirable characteristics in the proposed approach: 
(i) \emph{automatic learning of multi-task architectures} based on branching,
(ii) \emph{selective sharing} among tasks with automated learning of whom to share with.
In addition, we want our multi-task models to have low memory footprint and low latency during prediction (forward pass through the network). 

A natural approach for enabling sharing across multiple tasks is to share model parameters (partially or fully) across 
the corresponding layers of the task-specific deep neural networks. At an extreme, we can imagine 
a fully shared multi-task network architecture where all layers are shared except the last layer 
which predicts the labels for individual tasks. However, this unrestricted
sharing may suffer from the problem of \emph{negative transfer} where inadequate sharing across 
two unrelated tasks can worsen the performance on both. To avoid this, most of the multi-task
deep architectures share the bottom layers till some layer $l$ after which the sharing is blocked, 
resulting in task-specific sub-networks or branches beyond it \cite{HyperFace16,Brendan16,huang2015cross}. 
This is motivated by the observation made by several earlier works that bottom layers capture 
low level detailed features, which can be shared across multiple tasks, whereas top
layers capture features at a higher level of abstraction that are more task specific.
It can be further extended to a more general tree-like architecture, e.g., a smaller group of tasks can share 
parameters even after the first break-point at layer $l$ and breakup at a later layer.   
However, the space of such possible branching architectures is combinatorially large and current approaches 
largely make a decision based on limited manual exploration of this space, often biased by designer's perception of the 
relationship among different tasks \cite{Misra16}.

Our goal in this work is to develop a principled approach for designing multi-task deep learning
architectures obviating the need for tedious manual explorations. The proposed approach operates in a greedy
top-down manner, making branching and task-grouping decisions at each layer of the network using
a novel criterion that promotes the creation of separate branches for unrelated tasks (or groups of tasks) while 
penalizing for the model complexity. Since we also desire  a multi-task model with low memory 
footprint, the proposed approach starts with a \emph{thin} network and dynamically grows it
during the training phase by creating new branches based on the aforementioned criterion. 
We also propose a method based on simultaneous orthogonal matching pursuit (SOMP) \cite{somp}
for initializing a thin network from a pretrained wider network
(e.g., VGG-16) as a side contribution in this work.  

We evaluate the proposed approach on person attribute classification, where each attribute is considered a task (with non-mutually exclusive labels),
achieving state-of-the-art results with highly compact multi-task models.
On the CelebA dataset \cite{liu2015deep}, we match the current top results on facial attribute classification (90\% accuracy) with a model 90x more compact and 3x faster than the original VGG-16 model. We draw similar conclusions for clothing category recognition on the DeepFashion dataset \cite{liu2016deepfashion}, demonstrating that we can perform simultaneous facial and clothing attribute prediction using a single compact multi-task model, while preserving accuracy.

\noindent In summary, our main contributions are listed below:
\begin{compactitem}[$\circ$]
	\item We propose to automate learning of multi-task deep network architectures through a novel dynamic branching procedure,
	which makes task grouping decisions at each layer of the network (deciding with whom each task should share features) by 
	taking into account both task relatedness and complexity of the model.
	
	\item A novel method based on Simultaneous Orthogonal Matching Pursuit is proposed for initializing a thin network from a wider pre-trained network model,
	leading to faster convergence and higher accuracy.
	
	\item We perform {\em joint prediction} of facial and clothing attributes, achieving state-of-the-art results on standard datasets with a significantly 
	more compact and efficient multi-task model.  We also conduct relevant ablation studies providing insights into the proposed approach. 
\end{compactitem}

\section{Related Work}

{\bf Multi-Task Learning.} There is a long history of research in multi-task learning \cite{Caruana97,Thrun98,jacob2009clustered,Abhishek12,Misra16}. Most proposed techniques assume that all tasks are related and appropriate for joint training. A few methods have addressed the problem of ``with whom'' each task should share features \cite{xue2007multi,jacob2009clustered,zhou2011clustered,Kristen11,Abhishek12,passos2012flexible}. These methods are generally designed for shallow classification models, while our work investigates feature sharing among tasks in hierarchical models such as deep neural networks.

Recently, several methods have been proposed for multi-task learning using deep neural networks. HyperFace \cite{HyperFace16} simultaneously learns to perform face detection, landmarks localization, pose estimation and gender recognition. UberNet \cite{UberNet16} jointly learns low-, mid-, and high-level computer vision tasks using a compact network model.  MultiNet \cite{MultiNet16} exploits recurrent networks for transferring information across tasks. Cross-ResNet \cite{Brendan16} connects tasks through residual learning for knowledge transfer. However, all these methods rely on {\em hand-designed} network architectures composed of base layers that are shared across tasks and specialized branches that learn task-specific features.

As network architectures become deeper, defining the right level of feature sharing across tasks through handcrafted network branches is impractical. Cross-stitching networks \cite{Misra16} have been recently proposed to learn an optimal combination of shared and task-specific representations. Although cross-stitching units connecting task-specific sub-networks are designed to \emph{learn} the  feature sharing among tasks, 
the size of the network grows linearly with the number of tasks, causing scalability issues. We instead propose a novel algorithm that makes decisions about branching  based on task relatedness, while optimizing for the efficiency of the model. We note that other techniques such as HD-CNN \cite{HDCNN15} and Network of Experts \cite{ahmed2016network} also group related classes to perform hierarchical classification, but these methods are not applicable for the multi-label setting (where labels are not mutually exclusive).

{\bf Model Compression and Acceleration.} Existing deep convolutional neural network models are computationally and memory intensive, hindering their deployment in devices with low memory resources or in applications with strict latency requirements.  Methods for compressing and accelerating convolutional networks include knowledge distillation \cite{Hinton15,romero2014fitnets}, low-rank-factorization \cite{ioannou2015training,tai2015convolutional,sainath2013low}, pruning and quantization \cite{han2015deep,polyak2015channel}, structured matrices \cite{Circulant15,sindhwani2015structured,gong2016tamp}, and dynamic capacity networks \cite{almahairi2015dynamic}. These methods are task-agnostic and therefore most of them are complementary to our approach, which seeks to obtain a compact multi-task model by widening a low-capacity network based on task relatedness. 
Moreover, many of these state-of-the-art compression techniques can be used to further reduce the size of our learned multi-task architectures. 

{\bf Person Attribute Classification.} Methods for recognizing attributes of people, such as facial and clothing attributes, have received increased attention in the past few years. In the visual surveillance domain, person attributes serve as features for improving person re-identification \cite{su2016deep} and enable search of suspects based on their description \cite{vaquero2009attribute,feris2014attribute}. In e-commerce applications, these attributes have proven effective in improving clothing retrieval \cite{huang2015cross}, and fashion recommendation \cite{liu2014wow}. It has also been shown that facial attribute prediction is helpful as an auxiliary task for improving face detection \cite{yang2015facial} and face alignment \cite{zhang2016learning}.

State-of-the-art methods for person attribute prediction are based on deep convolutional neural networks \cite{wang2016walk,liu2015deep,chen2015deep,zhang2014panda}. Most methods either train separate classifiers per attribute \cite{zhang2014panda} or perform joint learning with a fully shared network \cite{rudd2016moon}.  Multi-task networks have been used with base layers that are shared across all attributes, and branches to encode task-specific features for each attribute category \cite{huang2015cross,sudowe2015person}. However, in contrast to our work, the network branches are hand-designed and do not exploit the fact that some attributes are more related than others in order to determine the level of sharing among tasks in the network. Moreover, we show that our approach produces a single compact network that can predict both facial and clothing attributes simultaneously.

\section{Methodology}

Let the linear operation in a layer $l$ of the network be paramterized by $W^l$.
Let $x^l \in \mathbb{R}^{c_l}$ be the input vector of layer $l$, and $y^l \in \mathbb{R}^{c_{l+1}}$ be the output vector. In feedforward networks that are of interest to this work, it is always the case that $x^l = y^{l-1}$. In other words, the output of a layer is the input to the layer above. In vision applications, the feature maps are often considered as three-way tensors and one should think of $x^l$ and $y^l$ as  appropriately vectorized versions of the input and output feature tensors. The functional form of the network is a series of within-layer computations chained in a sequence linking the lowest to the highest (output) layer. The within-layer computation (for both convolutional and fully-connected layers) can be concisely represented by a simple linear operation parametrized by $W^l$, followed by a non-linearity $\sigma_l(\cdot)$ as 
\begin{equation}
\label{eqn:within_layer}
y^l = \sigma_l(\mathcal{P}(W^l) x^l),
\end{equation}
\noindent where $\mathcal{P}$ is an operator that maps the parameters $W^l$ to the appropriate matrix $\mathcal{P}(W^l)\in\mathbb{R}^{c_{l+1}\times c_l}$.
For a fully connected layer $\mathcal{P}$ reduces to the identity operator, whereas for a convolutional layer with $f_l$ filters, $W^l\in\mathbb{R}^{f_l\times d_l}$ contains the vectorized filter coefficients in each row and the operator $\mathcal{P}$ maps it to an appropriate matrix that represents convolution as matrix multiplication. 
With this unified representation, we define the width of the network at layer $l$ as $c_l$ for the fully connected layers, and as $f_l$ for the convolutional layers. 
The widths at different layers are critical hyper-parameters for a network design. In general, a wider network is more expensive to train and deploy, but it has the capacity to model a richer set of visual patterns. The relative width across layers is a particularly relevant consideration in the design of a multi-task network. It is widely observed that higher layer represents a level of abstraction that is more task dependent. This is confirmed by previous works on visualization of filters at different layers \cite{zeiler2014visualizing}.  

Traditional approaches tackle the width design problem largely through hand-crafted layer design and manual model selection. Notably, popular deep convolutional network architectures, such as AlexNet \cite{krizhevsky2012imagenet}, VGG \cite{simonyan2014very}, Inception \cite{szegedy2015going} and ResNet \cite{he2015deep}
all use wider layers at the top of the network in what can be called an ``inverse pyramid'' pattern. These architectures serve as excellent reference designs in a myriad of domains, but researchers have noted that the width schedule (especially at the top layers) need to be tuned for the underlying set of tasks the network has to perform in order to achieve best accuracy \cite{Misra16}.

Here we propose an algorithm that dynamically finds the appropriate width of the multi-task network along with the task groupings through a multi-round training procedure. It has three main phases:

{\bf Thin Model Initialization.}
We start with a thin neural network model, initializing it from a pre-trained wider VGG-16 model by selecting a subset of filters using simultaneous orthogonal matching pursuit (ref. Section \ref{sec:somp}).   

{\bf Adaptive Model Widening.}
The thin initialized model goes through a multi-round widening and training procedure. The widening is done in a greedy top-down layer-wise manner starting from the top layer. For the current layer to be widened, our algorithm makes a decision on the number of branches to be created at this layer along with task assignments for each branch. The network architecture is frozen when the algorithm decides to create no further branches (ref. Section \ref{sec:widen}).   

{\bf Training with the Final Model.} In this last phase, the fixed final network is trained 
until convergence. 

More technical details are discussed in the next few sections. Algorithm \ref{alg:outline} provides a summary of the procedure. 

\begin{algorithm}[!t]\label{alg:outline} \small
	\SetInd{1ex}{1ex}
	\KwData{Input data $D=(x_i,y_i)_{i=1}^N$. The labels $y$ are for a set of $T$ tasks.}
	\KwIn{Branch factor $\alpha$, and thinness factor $\omega$. Optionally, a pre-trained network $M_p$ with parameters $\Theta_p$.}
	\KwResult{A trained network $M_f$ with parameters $\Theta_f$.}
	{\bf Initialization}: $M_0$ is a thin-$\omega$ model with $L$ layers. \\ 
	\eIf {exist $M_p,\Theta_p$} {
		$\Theta_0 \leftarrow \text{\em SompInit}(M_0, M_p, \Theta_p)$. $t \leftarrow 1$, $d \leftarrow T$. (Sec. \ref{sec:somp}) \\
	}{
	$\Theta_0 \leftarrow$ Random initialization
}
\While{($t \leq L$) and ($d > 1$)}{
	$\Theta_t, A_t\leftarrow \text{\em TrainAndGetAffinity}(D, M_t, \Theta_t)$ (Sec. \ref{sec:aff}) \\
	$d \leftarrow \text{\em FindNumberBranches}(M_t, A_t, \alpha)$ (Sec. \ref{sec:width_sel}) \\
	$M_{t+1}, \Theta_{t+1} \leftarrow \text{\em WidenModel}(M_t, \Theta_t, A_t, d)$ (Sec. \ref{sec:widen}) \\
	$t \leftarrow t+1$
}
Train model $M_t$ with sufficient iterations, update $\Theta_t$. 
$M_f \leftarrow M_t$, $\Theta_f \leftarrow \Theta_t$.  
\caption{Training with Adaptive Widening}
\end{algorithm}

\begin{figure}[t]
	\begin{center}
		\includegraphics[width=3.2in, height=1.2in]{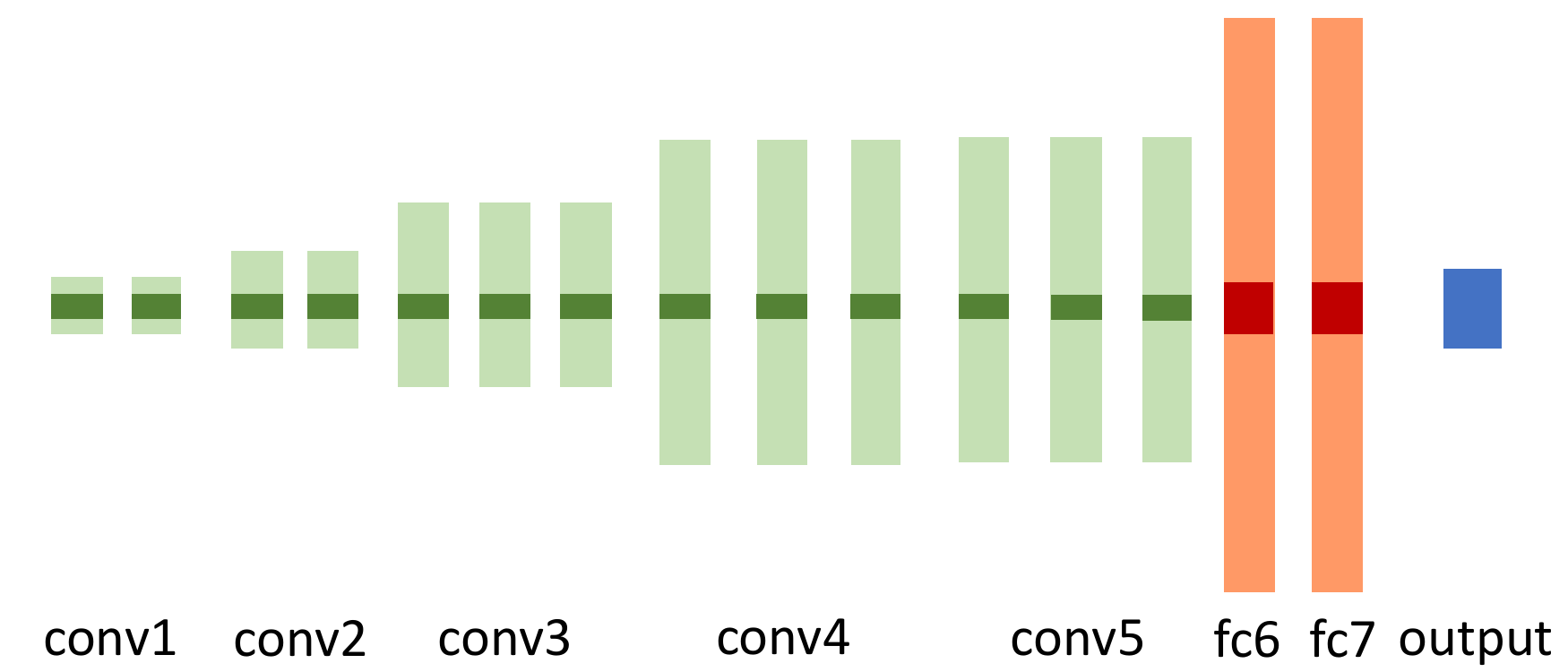}  
	\end{center}
	\caption{Comparing the thin model with VGG-16. As shown, the light color blobs shows the layers in the VGG-16 architecture. It has an inverse pyramid structure with a width plan of 64-128-256-512-512-4096-4096. The dark color blobs shows a thin network with $\omega=32$. The convolutional layers all have widths of $32$, and the fully connected layers have widths of $64$.}
	\label{fig:cmp_vgg16}
	\vspace{-2mm}
\end{figure}

\subsection{Thin Networks and Filter Selection using Simultaneous Orthogonal Matching Pursuit}
\label{sec:somp}
The initial model we use is a thin version of the VGG-16 network. It has the same structure as VGG-16 except for the widths at each layer. We experiment with a range of thin models that are denoted as thin-$\omega$ models. The width of a convolutional layer of the thin-$\omega$ model is the minimum between $\omega$ and the width of the corresponding layer of the VGG-16 network. The width of the fully connected layers are set to $2\omega$. We shall call $\omega$ the ``thinness factor''. Figure \ref{fig:cmp_vgg16} illustrates a thin model side by side with VGG-16. 

Using weights from pre-trained models is known to speed up training and improve model generalization. However, the standard direct copy method is only suitable when the source and the target networks have the same architecture (at least for most of the layers). Our adoption of a thin initial model forbids the use of direct copy, as there is a mismatch in the dimension of the weight matrix (for both the input and output dimensions, see Equation \ref{eqn:within_layer} and discussions). In the literature a set of general methods for training arbitrarily small networks using an existing larger network and the training data are known as ``knowledge distillation' \cite{Hinton15,romero2014fitnets}. However, for the limited use case of this work we propose a faster, data-free, and simple yet reasonably effective method. 
Let $W^{p,l}$ be the parameters of the pre-trained model at layer $l$ with $d$ rows. For convolutional layers, each row of $W^{p,l}$ represents a vectorized filter kernel. The initialization procedure aims to identify a subset of $d' (<d)$ rows of $W^{p,l}$ to form $W^{0,l}$ (the superscript $0$ denotes initialized parameters for the thin model). 
We would like the selected rows that minimize the following objective:
\begin{equation}
\label{eqn:somp}
A^\star, \omega^{\star}(l) = \underset{A \in \mathbb{R}^{d \times d'}, |\omega|=d'}{\arg\min} ||W^{p,l} - AW^{p,l}_{\omega:}||_F,
\vspace{-1mm}
\end{equation}
\noindent where $W^{p,l}_{\omega:}$ is a truncated weight matrix that only keeps the rows indexed by the set $\omega$. This problem is NP-hard, however, there exist approaches based on convex relaxation \cite{tropp2006algorithms} and greedy simultaneous orthogonal matching pursuit (SOMP) \cite{somp} which can produce approximate solutions. 
We use the greedy SOMP to find the approximate solution $\omega^{\star}(l)$ 
which is then used to initialize the parameter matrix of the thin model as $W^{0,l}\leftarrow W^{p,l}_{\omega^{\star}(l):}$.
We run this procedure layer by layer, starting from the input layer. At layer $l$, after initializing $W^{0,l}$, we replace $W^{p,l+1}$ with a column-truncated version that only keeps the columns indexed by $\omega^{\star}(l)$ to keep the input dimensions consistent. This initialization procedure is applicable for both convolutional and fully connected layers. See Algorithm \ref{alg:init}. 

\begin{algorithm}[!t]\label{alg:init} \small
	
	\KwIn{The architecture of the thin network $M_0$ with $L$ layers. The pretrained network and its parameters $M_p$, $\Theta_p$. Denote the weight matrix at layer $l$ as $W^{p,l} \in \Theta_p$.}
	\KwResult{The initial parmaeters of the thin network $\Theta_0$.}
	\ForEach{$l \in 1, 2, \cdots, L$}{
		Find $\omega^{\star}(l)$ in Equation \ref{eqn:somp} by SOMP, using $W^{p,l}$ as weight matrix. \\
		$W^{0,l} \leftarrow {W^{p,l}_{\omega^{\star}(l):}}$ \\
		$W^{p,l+1} \leftarrow \left((W^{p,l+1})^T_{\omega^{\star}(l):}\right)^T$
	}
	Aggregate $W^{0,l}$ for $l \in \{1, 2, \cdots, L\}$ to form $\Theta_0$. 
	\caption{SompInit($M_0$, $M_p$, $\Theta_p$)}
\end{algorithm}

\subsection{Top-Down Layer-wise Model Widening}
\label{sec:widen}

\begin{figure*}[t]
	\begin{center}
		\includegraphics[width=6.5in, height=1.3in]{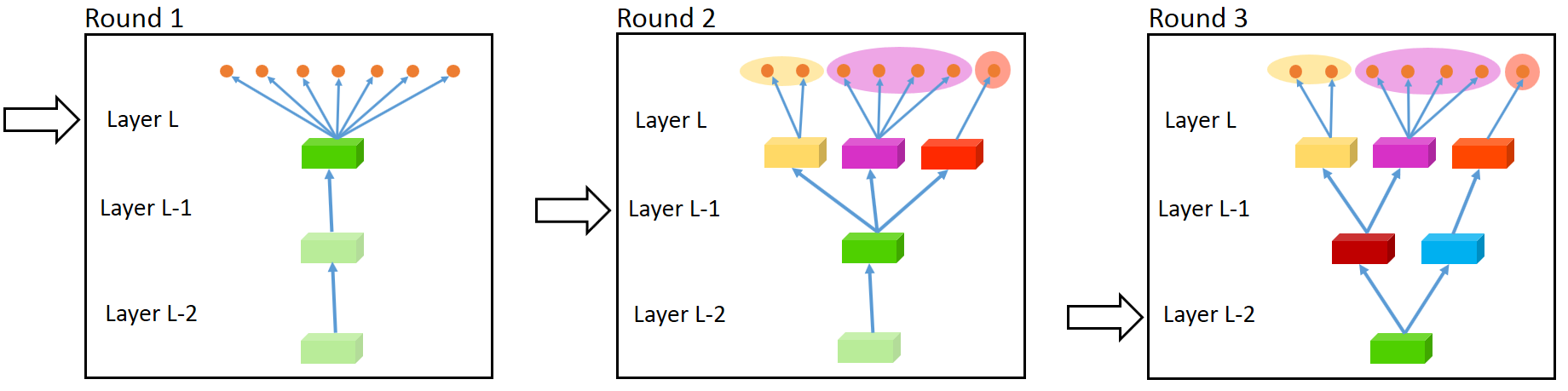}  
	\end{center}
	\caption{Illustration of the widening procedure. {\em Left:} the active layer is at layer $L$, there is one junction with 7 branches at the top. {\em Middle:} The seven branches are clustered into three groups. Three branches are created at layer $L$, resulting in a junction at layer $L-1$. Layer $L-1$ is now the active layer. {\em Right:}  Two branches are created at layer $L-1$, making layer $L-2$ now the active layer. At each branch creation, the filters at the newly created junction are initialized by direct copy from the old filter. }
	\label{fig:widen}
\end{figure*}

At the core of our training algorithm is a procedure that incrementally widens the current design in a layer-wise fashion. 
Let us 
introduce the concept of a ``junction''. A junction is a point at which the network splits into two or more independent sub-networks. We shall call such a sub-network a ``branch''. Each branch leads to a subset of prediction tasks performed by the full network. In the context of person attributes classification, each prediction is a {\em sigmoid} unit that produces a normalized confidence score on the existence of an attribute.

We propose to widen the network only at these junctions. More formally, consider a junction at layer $l$ with input $x^l$ and $d$ outputs $\{y^l_i\}_{i=1}^d$. Note that each output is the input to one of the $d$ top sub-networks. Similar to Equation \ref{eqn:within_layer} the within-layer computation is given as
\begin{equation}
\vspace{-1mm}
\label{eqn:widen_high}
y^l_i = \sigma_l(\mathcal{P}(W^l_i) x^l)  \quad\quad \mbox{for} \quad i \in [d],
\end{equation}
\noindent where $W^l_i$ parameterizes the connection from input $x^l$ to the $i$'th output $y^l_i$ at layer $l$. 
The set $[d]$ is the indexing set $\{1, 2, \cdots, d\}$. A junction is widened by creating new outputs at the layer below. To widen layer $l$ by a factor of $c$, we make layer $l-1$ a junction with $2 \leq c \leq d$ outputs. We use $y^{l-1}_j$ to denote an output in layer $l-1$ (each is an input for layer $l$) and $W^{l-1}_j$ to denote its parameter matrix. All of the newly-created parameter matrices have the same shape as $W^{l-1}$ (the parameter matrix before widening). The single output $y^{l-1}=x^l$ is replaced by a set of outputs $\{y^{l-1}_j\}_{j=1}^{c}$ where
\begin{equation}
\label{eqn:widen_low}
y_j^{l-1} = \sigma_{l-1} (\mathcal{P}(W^{l-1}_j) x^{l-1}) \quad\quad \mbox{for} \quad j \in [c].
\end{equation}
Let $g^l : [d] \to [c]$ be a given grouping function at layer $l$. After widening, the within-layer computation at layer $l$ is given as (cf. Equation \ref{eqn:widen_high})
\begin{equation}
\label{eqn:widened}
y_i^l = \sigma_l(\mathcal{P}(W^l_i) x^l_{g^l(i)}) = \sigma_{l}\left(\mathcal{P}(W^l_i) \sigma_{l-1}(\mathcal{P}(W^{l-1}_{g^l(i)}) x^{l-1})\right)
\end{equation}
\noindent where the latter equality is a consequence of Equation \ref{eqn:widen_high}. The widening operation sets the initial weight for $W^{l-1}_j$ to be equal to the original weight of $W^{l-1}$. It allows the widened network to preserve the functional form of the smaller network, enabling faster training.

To put the widening of one junction into the context of the multi-round progressive model widening procedure, consider a situation where there are $T$ tasks. Before any widening, the output layer of the initial thin multi-task network has a junction with $T$ outputs, each is the output of a sub-network (branch). It is also the only junction at initialization. The widening operation naturally starts from the output layer (denoted as layer $l$). It will cluster the $T$ branches into $t$ groups where $t \leq T$. In this manner the widening operation creates $t$ branches at layer $l-1$. The operation is performed recursively in a top-down manner towards the lower layers. Note that each branch will be associated with a sub-set of tasks. There is a 1-1 correspondence between tasks and branches at the output layer, but the granularity goes coarser at lower layers. An illustration of this procedure can be found in Figure \ref{fig:widen}. 

\subsection{Task Grouping based on the Probability of Concurrently Simple or Difficult Examples}
\label{sec:aff}

Ideally, dissimilar tasks are separated starting from a low layer, resulting in less sharing of features. For similar tasks the situation is the opposite. We observe that if an easy example for one task is typically a difficult example for another, intuitively a distinctive set of filters are required for each task to accurately model both in a single network. Thus we define the affinity between a pair of tasks as the probability of observing concurrently simple or difficult examples for the underlying pair of tasks from a random sample of the training data. 

To make it mathematically concrete, we need to properly define the notion of a ``difficult'' and a ``simple'' example. Consider an arbitrary attribute classification task $i$. Denote the prediction of the task for example $n$ as $s_i^n$, and the error margin as $m_i^n = |t_i^n - s_i^n|$, where $t_i^n$ is the binary label for task $i$ at sample $n$. Following the previous discussion, it seems natural to set a fixed threshold on $m_i^n$ to decide whether example $n$ is simple or difficult. However, we observe that this is problematic since as the training progresses most of the examples will become simple as the error rate decreases, rendering this measure of affinity useless. An adaptive but universal (across all tasks) threshold is also problematic as it creates a bias that makes intrinsically easier tasks less related to all the other tasks. 

These observations lead us to the following approach. Instead of setting a fixed threshold, we estimate the average margin for each task, 
$\mathbb{E}\{m_i\}$. We define the indicator variable for a difficult example for task $i$ as $e_i^n = \mathbf{1}_{m_i^n \geq \mathbb{E}\{m_i\}}$.
For a pair of tasks $i$, $j$, we define their affinity as 
\begin{eqnarray}
\vspace{-1mm}
\label{eqn:aff}
A(i,j) & = & \mathbb{P}(e_i^n=1, e_j^n=1) + \mathbb{P}(e_i^n=0, e_j^n=0) \nonumber \\
& = & \mathbb{E}\{e_i^n e_j^n + (1-e_i^n)(1-e_j^n)\}.
\vspace{-1mm}
\end{eqnarray}
Both $\mathbb{E}\{m_i\}$ and the expectation on Equation \ref{eqn:aff} can be estimated by their sample averages. Since these expectations are functions of the current neural network model, a naive implementation would require a large number of time consuming forward passes after every training iterations. As a much more efficient implementation, we alternatively collect the sample averages from each training mini-batches. The expectations are estimated by computing a weighted average of the within-batch sample averages. To make the estimation closer to the true expectations from the current model, an exponentially decaying weight is used. 

The estimated task affinity is used directly for the clustering at the output layer. It is natural as branches at the output layer has a 1-1 map to the tasks. But at lower layers the mapping is one to many, as a branch can be associated with more than one tasks. In this case, affinity is computed to reflect groups of tasks. In particular, let $k$, $l$ denote two branches at the current layer, where $i_{k}$ and $j_{l}$ denotes the $i$-th and $j$-th task associated with each branch respectively. The affinity of the two branches are defined by
\begin{eqnarray}
\vspace{-1mm}
\label{eqn:between_sep}
\tilde{A_b}(k, l) &=& \underset{i_k}{\mbox{mean}}\left(\underset{j_l}{\min}\quad {A(i_{k}, j_{l})}\right) \\
\tilde{A_b}(l, k) &=& \underset{j_l}{\mbox{mean}}\left(\underset{i_k}{\min}\quad {A(i_{k}, j_{l})}\right)
\vspace{-1mm}
\end{eqnarray}
The final affinity score is computed as $A_b(k, l) = (\tilde{A_b}(k, l) + \tilde{A_b}(l, k))/2$.
Note that if branches and tasks form a 1-1 map (the situation at the output layer), 
this
reduces to the definition in Equation \ref{eqn:aff}. For branches with coarser task granularity, 
$A_b(k, l)$
measures the affinity between two branches by looking at the largest distance (smallest affinity) between their associated tasks.

\subsection{Complexity-aware Width Selection}
\label{sec:width_sel}

The number of branches to be created determines how much wider the network becomes after a widening operation. This number is determined by a loss function that balances complexity and the separation of dissimilar tasks to different branches. For each number of clusters $1\leq d \leq c$, we perform spectral clustering to get a grouping function $g_d: [d] \to [c]$ that associates the newly created branches with the $c$ old branches at one layer above. At layer $l$ the loss function is given by
\begin{equation}
\label{eqn:cost_widen}
L^l(g_d) = (d-1) L_0 2^{p_l} + \alpha L_s(g_d)
\end{equation}
\noindent where $(d-1) L_0 2^{p_l}$ is a penalty term for creating branches at layer $l$, $L_s(g_d)$ is a penalty for separation. $p_l$ is defined as the number of pooling layers above the layer $l$ and $L_0$ is the unit cost for branch creation. The first term grows linearly with the number of branches, with a scalar that defines how expensive it is to create a branch at the current layer (which is heuristically set to double after every pooling layers). Note that in this formulation a larger $\alpha$ encourages the creation of more branches. We call $\alpha$ the branching factor. The network is widened by creating the number of branches that minimizes the loss function, or ${g_d^{l}}^{\star} = \underset{g_d}{\arg\min}\quad {L^l(g_d)}$.

The separation term is a function of the branch affinity matrix $A_b$. For each $i \in [d]$, we have
\begin{equation}
\vspace{-2mm}
\label{eqn:within_sep}
L_s^{i}(g_d) = 1 - \underset{k \in g^{-1}(i)}{\mbox{mean}}\left(\underset{l \in g^{-1}(i) }{\min}{A_b(k,l)}\right),
\end{equation}
\noindent and the separation cost is the average across each newly created branches
\begin{equation}
\vspace{-2mm}
L_s(g_d) = \frac{1}{d} \underset{i \in [d]}{\sum} L_s^i(g_d).
\end{equation}

Note Equation \ref{eqn:within_sep} measures the maximum distances (minimum affinity) between the tasks within the same group. It penalizes cases where very dissimilar tasks are included in the same branch. 

\section{Experiments}

\begin{table*}
	\begin{center} \small
		\begin{tabular}{|l|c|c|c|c|c|}
			\hline
			\bf Method & \bf Accuracy (\%) & \bf Top-10 Recall (\%) & \bf Test Speed (ms) & \bf Parameters (millions) & \bf Jointly? \\ \hline
			LNet+ANet & 87 & N/A & $+$ & $+$ & No \\
			Walk and Learn & 88 & N/A & $+$ & $+$ & No \\
			MOON & 90.94 & N/A & $\approx 33^*$ & 119.73 & No \\ \hline
			Our VGG-16 Baseline & 91.44 & 73.55 & 33.2 & 134.41 & No \\
			Our Low-rank Baseline & 90.88 & 69.82 & 16.0 & 4.52 & No \\
			Our Baseline-thin-32 & 89.96 & 65.95 & 5.1 & 0.22 & No \\ \hline
			Our Branch-32-1.0 & 90.74 & 69.95 & 9.6 & 1.49 & No \\
			Our Branch-32-2.0 & 90.90 & 71.08 & 15.7 & 2.09 & No \\
			Our Branch-64-1.0 & 91.26 & 72.03 & 15.2 & 4.99 & No \\ \hline
			Our Joint Branch-32-2.0 & 90.4 & 68.72 & 10.01 & 3.25 & Yes \\
			Out Joint Branch-64-2.0 & 91.02 & 71.38 & 16.28 & 10.53 & Yes \\
			\hline
		\end{tabular}
	\end{center}
	\caption{Comparison of accuracy, speed and compactness on CelebA test set. LNet+ANet and Walk and Learn results are cited from \cite{wang2016walk}. MOON results are cited from \cite{rudd2016moon}. $+$: There is no reported number to cite. $*$: MOON uses the VGG16 architecture, thus its test time should be similar to our VGG-16 baseline.}
	\label{tab:celeba_complexity}
\end{table*}

\begin{table*}
	\begin{center} \small
		\begin{tabular}{|l|c|c|c|c|c|}
			\hline
			\bf Method & \bf Top-3 Accuracy (\%) & \bf Top-5 Accuracy (\%) & \bf Test Speed (ms) & \bf Parameters (millions) & \bf Jointly? \\ \hline
			WTBI & 43.73 & 66.26 & $+$ & $+$ & No \\
			DARN & 59.48 & 79.58 & $+$ & $+$ & No \\
			FashionNet & ${82.58}^{\#}$ & ${90.17}^{\#}$ & $\approx 34^*$ & $\approx 134^*$ & No \\ \hline
			Our VGG-16 Baseline & 86.72 & 92.51 & 34.0 & 134.45 & No \\
			Our Low-rank Baseline & 84.14 & 90.96 & 16.34 & 4.52 & No \\ \hline
			Our Joint Branch-32-2.0 & 79.91 & 88.09 & 10.01 & 3.25 & Yes \\
			Our Joint Branch-64-2.0 & 83.24 & 90.39 & 16.28 & 10.53 & Yes \\
			\hline
		\end{tabular}
	\end{center}
	\caption{Comparison of accuracy, speed and compactness on Deepfashion test set. WTBI and DARN results are cited from \cite{liu2016deepfashion}. The experiments are reportedly performed in the same condition on the FashionNet method and tested on the DeepFashion test set. $+$: There is no reported number to cite. $*$: There is no reported number, but based on the adoption of VGG-16 network as base architecture they should be similar to those of our VGG-16 baseline. $\#$: The results are from a network jointly trained for clothing landmark, clothing attribute and clothing categories predictions. We cite the reported results for clothing category \cite{liu2016deepfashion}.}
	\label{tab:deepfashion_complexity}
	\vspace{-3mm}
\end{table*}

We perform an extensive evaluation of our approach on person attribute classification tasks. We use CelebA \cite{liu2015deep} dataset for facial attribute classification tasks and Deepfashion \cite{liu2016deepfashion} for clothing category classification tasks. 
CelebA consists of images of celebrities labeled with 40 attribute classes. Most images also include the torso region in addition to the face. Our models are evaluated using the standard classification accuracy (average of classification accuracy rate over all attribute classes) and the top-10 recall rate (proportion of correctly retrieved attributes from the top-10 prediction scores for each image). Top-10 is used as there are on average about 9 positive facial attributes per image on this dataset. 
DeepFashion is richly labeled with 50 categories of clothes, such as ``shorts'', ``jeans'', ``coats'', etc. (the labels are mutually exclusive).  Faces are often visible on these images. We evaluate top-3 and top-5 classification accuracy to directly compare with benchmark results in \cite{liu2016deepfashion}.


\subsection{Comparison with the State of the art}

We establish three baselines. The first baseline is a VGG-16 model initialized from the a model trained from imdb-wiki gender classification \cite{imdb}. The second baseline is a low-rank model with low rank factorization at all layers. This model is also initialized from the imdb-wiki gender pretrained model, but the initialization is through truncated Singular Value Decomposition (SVD) \cite{denton2014exploiting}. The number of basis filters is 8-16-32-64-64 for the convolutional layers, 64-64 for the two fully-connected layers and 16 for the output layer. The third is a thin model initialized using the SOMP initialization method introduced in Section \ref{sec:somp}, using the same pre-trained model. Our VGG-16 baselines are stronger than all previously reported methods, while the low-rank baselines closely matches the state-of-the-art while being faster and more compact. The thin baseline is up to 6 times faster, 500 times more compact than the VGG-16 baseline, but still reasonably accurate. 

We find several contributing factors to the strength of our baselines. Firstly, the choice of pre-trained model is critical. Most recent works use the VGG face descriptor, whereas in our work we use the pre-trained model from  imdb-wiki \cite{imdb-wiki}. For the thin baseline, it is also important to use Batch Normalization (BN) \cite{ioffe2015batch}. Without the adoption of BN layers the training error ceases to decrease after a small number of training iterations. We observe this phenomenon in both random initialization and SOMP initialization.

A comparison of the models generated by our adaptive widening algorithm with baseline results are shown in Table \ref{tab:celeba_complexity} and \ref{tab:deepfashion_complexity}. Our ``branching'' models achieves similar or better accuracy compared to these state-of-the-art methods, while being much more compact and faster. 

\subsection{Cross-domain Training of Joint Person Attribute Network}
\vspace{-1mm}
To examine the ability of our approach in handling cross-domain tasks, we train a network that jointly predict facial and clothing attributes. The model is trained on the union of the two training sets. Note that the CelebA dataset is not annotated with clothing labels, and the Deepfashion dataset is not annotated with facial attribute labels. To augment the annotations for both datasets, we use the predictions provided by the baseline VGG-16 models as soft training targets. We demonstrate that the joint model is comparable to the state-of-the-art on both facial and clothing tasks, while being a much more efficient combined model rather than two separate models. The comparison between the joint models with the baselines is shown in Table \ref{tab:celeba_complexity} and \ref{tab:deepfashion_complexity}.

\begin{figure*}[t]
	\begin{center}
		\includegraphics[width=6.8in]{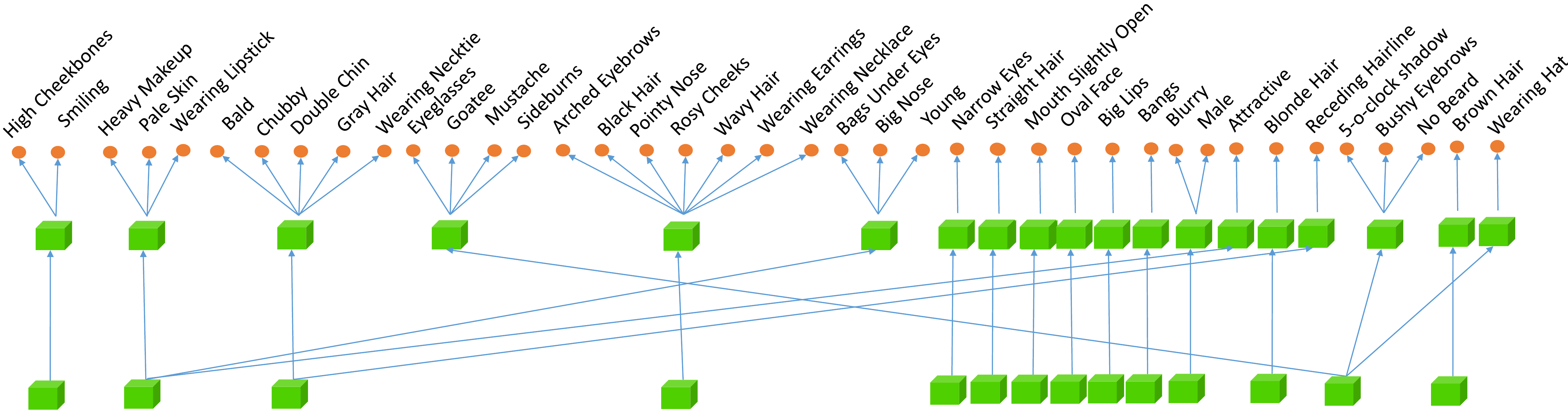}  
	\end{center}
	\caption{The actual task grouping in the Branch-32-2.0 model on CelebA. Upper: fc7 layer. Lower: fc6 layer. Other layers are omitted.}
	\label{fig:visual_grouping}
\end{figure*}

\vspace{-1mm}
\subsection{Visual Validation of Task Grouping}
We visually inspect the task groupings in the generated model. Figure \ref{fig:visual_grouping} displays the actual task grouping in the Branch-32-2.0 model trained on CelebA. The grouping are often highly intuitive. For instance, ``5-o-clock Shadow'', ``Bushy Eyebrows'' and ``No Beard'', which all describe some forms of facial hairs, are grouped. The cluster with ``Heavy Makeup'', ``Pale Skin'' and ``Wearing Lipstick'' is clearly related. Groupings at lower layers are also sensible. As an example, the group ``Bags Under Eyes'', ``Big Nose'' and ``Young'' are joined by ``Attractive'' and ``Receding Hairline'' at fc6, probably because they all describe age cues. This is particularly interesting as no human intervention is involved in model generation. 

\begin{figure}[t]
	\begin{center}
		\includegraphics[width=3.6in]{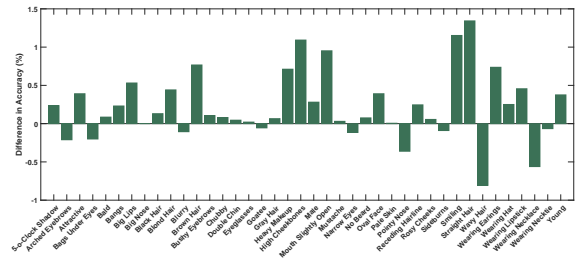}  
	\end{center}
	\caption{The reduction in accuracy when changing the task grouping to favor grouping of dissimilar tasks. A positive number suggests a reduction in accuracy when changing from original to the new grouping. This figure shows our automatic grouping strategy improves accuracy for most tasks. }
	\label{fig:task_group}
\end{figure}

\vspace{-1mm}
\subsection{Ablation Studies}
\vspace{-1mm}
{\bf What are the advantages of grouping similar tasks?} We shuffle the correspondence between training targets and the output of the network for ``Branch-32-2.0'' model from CelebA and report the reduction in accuracies for each tasks. Both random and manual shuffling are tested but we only report the one from manual shuffling as they are similar. In particular, for manual shuffling we choose a new grouping of tasks so that the network separates many tasks that are originally in the same branch. Figure \ref{fig:task_group} summarizes our findings. Clearly grouping tasks according to similarity improves accuracy for most tasks. 

Closer examination yields other interesting observations. The three tasks that actually benefit from the shuffling significantly (unlike most of the tasks), namely ``wavy hair'', ``wearing necklace'' and ``pointy nose'' are all from the branch with the largest number of tasks. This is sensible as after the shuffling they are not forced to share filters with many other tasks. But other tasks from the same branch, namely ``black hair'' and ``wearing earrings'' are significantly improved from the original grouping. One possible explanation is that while grouping similar tasks allow them to benefit from multi-task learning, some tasks are intrinsically more difficult and require a wider branch. Our current design lacks the ability to change the width of a branch, which is an interesting future direction. 

\begin{table}
	\begin{center} \small
		\begin{tabular}{|l|c|c|}
			\hline
			\bf Method & \bf Accuracy (\%) & \bf Top-10 Recall (\%) \\ \hline
			w/ pre-trained & -0.54 &  -2.47 \\
			w/o pre-trained & -0.65 & -3.77  \\
			\hline
		\end{tabular}
	\end{center}
	\caption{Accuracy gap with and without initialization from pre-trained model, defined as accuracy of Branch-32-2.0 minus the one from VGG-16 Baseline.}
	\label{tab:capacity}
	\vspace{-4mm}
\end{table}

{\bf Sub-optimal use of pretrained network or smaller capacity?} The gap in accuracy between Branch-32-2.0 and VGG-16 baseline can be caused by sub-optimal use of the pretrained model or the intrinsically smaller capacity of the former. To determine if both factors contribute to the gap, we compare training the Branch-32-2.0 model and VGG-16 from scratch on CelebA. As neither model benefit from the information from a pre-trained network, we expect a much smaller gap in accuracy if the sub-optimal use of the pretrained model is the main cause. Our results summarized in Table \ref{tab:capacity} suggest that the smaller capacity of the Branch-32-2.0 model is likely the main reason for the accuracy gap.

\begin{figure}[t]
	\begin{center}
		\includegraphics[width=3.0in]{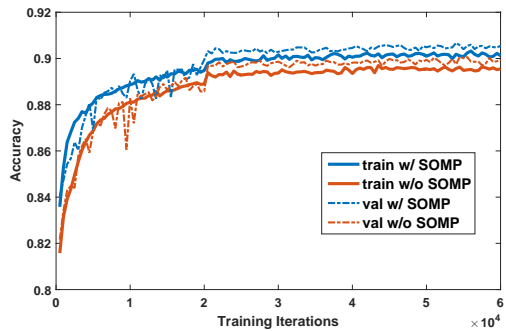}  
	\end{center}
	\caption{Comparison of training progress with and without SOMP initialization. The model using SOMP initialization clearly converges faster and better. }
	\label{fig:somp}
	\vspace{-4mm}
\end{figure}

{\bf How does SOMP help the training?} We compare training with and without this initialization using the Baseline-thin-32 model on CelebA, under identical training conditions. The evolution of training and validation accuracies are shown in Figure \ref{fig:somp}. Clearly, the network initialized with SOMP initialization converges faster and better than the one without SOMP initialization.
\vspace{-1mm}
\section{Conclusion}
\vspace{-1mm}
We have proposed a novel method for learning the structure of compact multi-task deep neural networks. Our method starts with a thin network model and expands it during training by means of a novel multi-round branching mechanism, which determines with whom each task shares features in each layer of the network, while penalizing for the complexity of the model. We demonstrated compelling results of the proposed approach on the problem of person attribute classification. As future work, we plan to adapt our approach to other related problems, such as incremental learning and domain adaptation.

{\small
	\bibliographystyle{ieee}
	\bibliography{egbib}
}

\end{document}